\documentclass[conference]{IEEEtran}
\IEEEoverridecommandlockouts
\usepackage{cite}
\usepackage{amsmath,amssymb,amsfonts}
\usepackage{algorithmic}
\usepackage{booktabs}

\usepackage{graphicx}
\usepackage{textcomp}
\usepackage{xcolor}
\def\BibTeX{{\rm B\kern-.05em{\sc i\kern-.025em b}\kern-.08em
    T\kern-.1667em\lower.7ex\hbox{E}\kern-.125emX}}
\begin{document}

\title{Depth-Copy-Paste: Multimodal and Depth-Aware Compositing for Robust Face Detection}


\author{%
  Qiushi Guo\\
  Coffee AI Lab, Great Wall Motor(GWM), China\\
  \texttt{guoqiushi910@gmail.com}\\

}

\maketitle

\begin{abstract}
Data augmentation is crucial for improving the robustness of face detection systems, especially under challenging conditions such as occlusion, illumination variation, and complex environments. Traditional copy–paste augmentation often produces unrealistic composites due to inaccurate foreground extraction, inconsistent scene geometry, and mismatched background semantics. To address these limitations, we propose Depth-Copy-Paste, a multimodal and depth-aware augmentation framework that generates diverse and physically consistent face detection training samples by copying full-body person instances and pasting them into semantically compatible scenes. Our approach first employs BLIP and CLIP to jointly assess semantic and visual coherence, enabling automatic retrieval of the most suitable background images for the given foreground person. To ensure high-quality foreground masks that preserve facial details, we integrate SAM3 for precise segmentation and Depth-Anything to extract only the non-occluded visible person regions, preventing corrupted facial textures from being used in augmentation. For geometric realism, we introduce a depth-guided sliding-window placement mechanism that searches over the background depth map to identify paste locations with optimal depth continuity and scale alignment. The resulting composites exhibit natural depth relationships and improved visual plausibility. Extensive experiments show that Depth-Copy-Paste provides more diverse and realistic training data, leading to significant performance improvements in downstream face detection tasks compared with traditional copy–paste and depth-free augmentation methods.
\end{abstract}

\section{Introduction}
Data augmentation plays a central role in training robust vision models, particularly for detection and segmentation tasks where collecting diverse real-world data is costly and time-consuming. Copy–paste augmentation\cite{ghiasi2021simple} has emerged as an effective strategy for enriching datasets by recombining foreground objects with new backgrounds. However, existing copy–paste pipelines often suffer from semantic mismatch, occlusion artifacts, and geometric inconsistencies, which ultimately limit their effectiveness in generating realistic training samples. In person-centric scenes, where human context, depth ordering, and visibility are critical, these shortcomings become even more pronounced.

A key difficulty lies in selecting backgrounds that are semantically compatible with a given foreground. Traditional augmentation either randomly chooses backgrounds or relies on simple heuristics such as scene class or coarse metadata. These approaches ignore deeper semantic cues e.g., activity context, lighting, or environment type leading to implausible composites. Meanwhile, foreground extraction methods frequently overlook occlusion patterns: many segmentation models correctly delineate human silhouettes but fail to distinguish occluded and visible body regions, causing artifacts when occluded regions are pasted into a new scene. Finally, geometric placement remains underexplored. Without reliable depth reasoning, pasted objects often float, overlap incorrectly with existing structures, or are mis-scaled relative to the background.
\begin{figure}[t]
    \centering
    \includegraphics[width=0.5\textwidth]{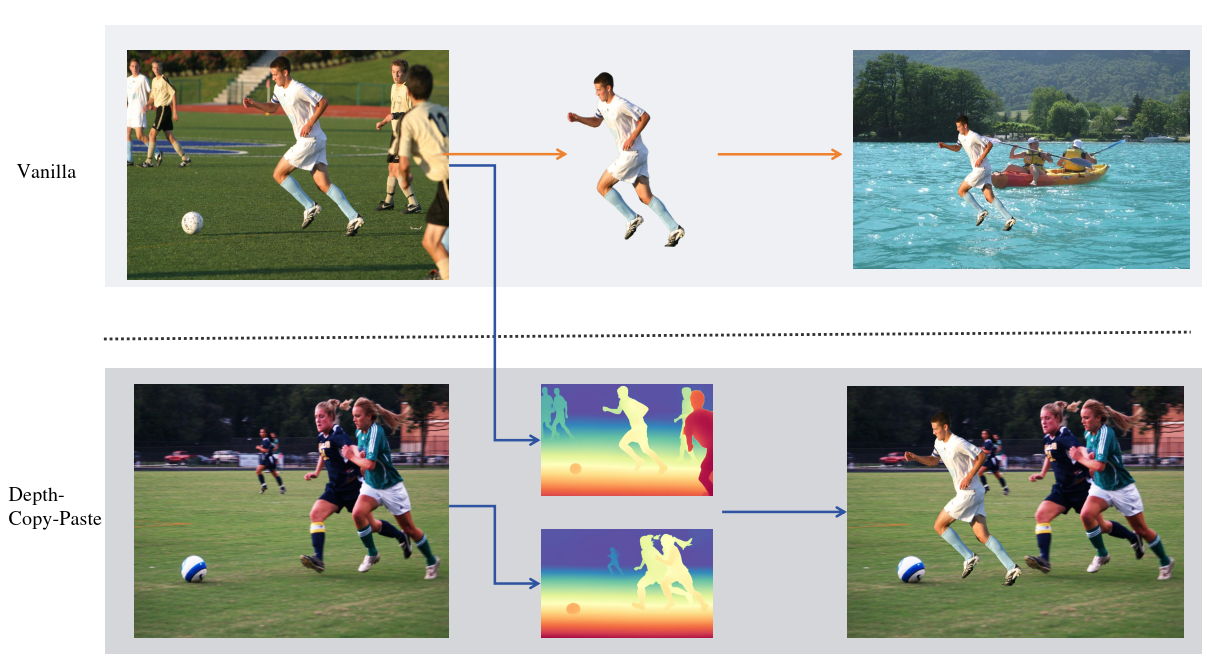}
    \caption{Comparison between vanilla copy–paste augmentation and our proposed Depth-Copy-Paste framework.
Top: Vanilla copy–paste directly segments and pastes the foreground person onto a new background without considering geometric consistency, often yielding unrealistic composites (e.g., incorrect scale or floating artifacts).
Bottom: Depth-Copy-Paste leverages depth estimation for both foreground and background to obtain non-occluded person masks and depth-aware placement. By matching depth distributions and enforcing geometric consistency, our method produces realistic and seamlessly integrated composites that better preserve scene plausibility.}
    \label{intro}
\end{figure}
To address these limitations, we propose \textit{Depth-Copy-Paste}, a multimodal and depth-aware compositing framework that generates high-fidelity synthetic images to enhance face detection models. Unlike traditional copy–paste augmentation, which often introduces semantic mismatch or geometric inconsistency, our approach jointly leverages semantic understanding and 3D reasoning to ensure that synthesized scenes remain realistic and contextually coherent. Specifically, we employ BLIP~\cite{li2022blip} and CLIP\cite{clip} to assess multimodal consistency between a foreground face region and a large pool of background candidates. This enables the retrieval of scenes that align with both textual cues and visual appearance, effectively reducing the domain gap that typically arises when foreground and background originate from heterogeneous sources.

To further ensure compositional realism, Depth-Copy-Paste integrates high-quality segmentation and geometric alignment into a unified pipeline. SAM3\cite{sam3} is used to obtain precise person masks, while depth estimates from Depth-Anything\cite{depthanything2024} allow us to isolate non-occluded and depth-consistent foreground regions, avoiding artifacts commonly observed in naive segmentation-based augmentation. Additionally, we introduce a depth-guided sliding-window placement strategy that selects background regions with compatible depth structure and scale, enabling the pasted person to blend naturally into the scene. Together, these multimodal retrieval, depth-aware extraction, and geometry-guided placement components form a coherent synthesis framework that substantially improves data diversity and boosts downstream face detection robustness.

Our contributions are threefold:
\begin{itemize}
    \item Multimodal Background Selection: We combine BLIP textual descriptions and CLIP visual–textual embeddings to retrieve semantically aligned backgrounds for each person foreground.
    \item Depth-Aware Foreground Extraction: We employ SAM3 segmentation with Depth-Anything depth estimation to isolate non-occluded person regions, avoiding artifacts from hidden surfaces.
    \item Extensive experiments show that Depth-Copy-Paste produces visually coherent composites and improves performance on downstream person detection tasks compared to traditional copy–paste and depth-agnostic baselines. Our results highlight the importance of integrating multimodal semantic cues and depth reasoning for high-quality synthetic data generation.
\end{itemize}

\section{Related Work}
\subsection{Copy-Paste}
Copy–paste augmentation has become an effective strategy for enriching training data in detection and segmentation tasks. Early methods typically perform random copy–paste, where object instances are extracted and pasted onto arbitrary backgrounds to increase appearance diversity, though such naive operations often introduce unrealistic artifacts due to missing contextual constraints. To address this, context-aware copy–paste approaches incorporate semantic cues, geometric relationships, or depth and perspective reasoning to ensure more plausible placement and better alignment with scene structure, leading to substantial gains in tasks sensitive to object–background interactions. In parallel, regional perturbation techniques such as Cutout, which masks random patches, and CutMix, which replaces image regions with those sampled from other images and mixes corresponding labels, provide strong regularization benefits but do not preserve instance-level structure; hence, they are less suited for modeling object compositionality. Overall, copy–paste based augmentation offers a more explicit mechanism for instance-level synthesis compared with traditional methods, enabling richer scene diversity and improved robustness.
\subsection{Face Depth}
Face detection has progressed significantly from early hand-crafted approaches to modern deep learning–based detectors optimized for large-scale benchmarks such as WIDER Face. Classical methods relied on Haar cascades\cite{haar}, HOG features\cite{hog}, and Deformable Part Models\cite{girshick2015deformable}, but their limited robustness under occlusion, scale variation, and illumination changes constrained deployment in unconstrained settings. With the emergence of CNNs, anchor-based detectors such as Faster R-CNN, SSD\cite{liu2016ssd}, and RetinaFace\cite{retinaface}, as well as lightweight variants like BlazeFace\cite{blazeface} and DBFace\cite{kim2023dcface}, achieved substantial improvements through multi-scale feature fusion, dense anchors, and dedicated facial priors. More recent anchor-free frameworks further simplified prediction while enhancing localization precision, especially in crowded or small-face scenarios. Despite these advances, robust face detection in the wild remains challenging due to heavy occlusion, extreme pose, and complex background context, motivating research on improved data augmentation, context modeling, and synthetic data generation to mitigate distribution gaps and enhance generalization.
\subsection{Depth Estimation}
Depth estimation has been a long-standing problem in computer vision, providing fundamental geometric cues for 3D scene understanding, reconstruction, and high-level perception tasks. Traditional approaches rely on stereo matching or multi-view geometry, where depth is inferred through epipolar constraints and hand-crafted correspondence features~\cite{scharstein2002taxonomy,hirschmuller2007stereo}. However, such methods often struggle in textureless regions, occlusions, and large-baseline configurations.

With the rise of deep learning, monocular depth estimation has become the mainstream direction. Early CNN-based frameworks, such as the seminal work by Eigen \textit{et al.}\cite{eigen2014depth}, introduced coarse-to-fine depth regression networks. Subsequent developments explored encoder--decoder architectures, scale-invariant losses, and spatial propagation modules\cite{laina2016deeper,hu2019revisiting}. More recently, transformer-based models and large-scale training pipelines, e.g., DPT\cite{ranftl2021vision}, MiDaS\cite{ranftl2020towards}, and Depth-Anything~\cite{depthanything2024} have significantly improved robustness and cross domain generalization through massive heterogeneous datasets and strong global-reasoning backbones.

In parallel, self-supervised depth estimation has emerged as a practical alternative when ground-truth depth is unavailable. Methods such as Monodepth\cite{godard2017unsupervised} and its successors\cite{godard2019digging,zhou2017unsupervised} use photometric consistency across video frames or stereo pairs as supervisory signals, enabling effective training without costly annotations.


\section{Method}
\begin{figure*}[htb]
    \centering
    \includegraphics[width=0.95\textwidth]{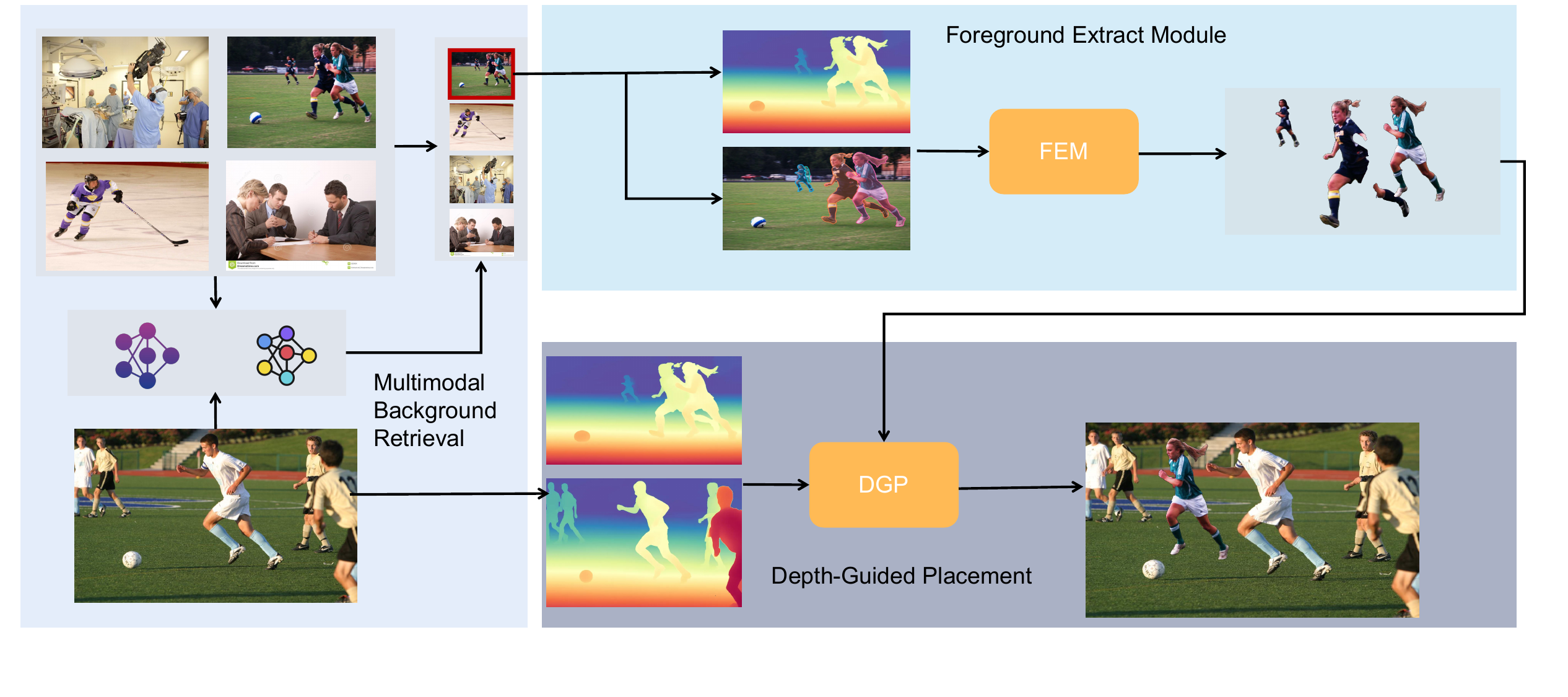}
    \caption{Overview of the proposed pipeline. (1) Multimodal Background Retrieval: BLIP-based captioning and CLIP-based visual–textual matching retrieve semantically aligned background scenes. (2) Foreground Extraction Module (FEM) integrates SAM3 segmentation with depth estimation to obtain clean and non-occluded masks. (3) Depth-Guided Placement (DGP) aligns normalized depth distributions and uses a sliding-window strategy to identify geometrically compatible placement regions. }
    \label{pipeline}
\end{figure*}
In this section, we present our Depth-Copy-Paste framework, which aims to generate geometrically consistent and semantically coherent composite face images. Unlike traditional copy--paste augmentation that performs foreground extraction and placement in a purely appearance-driven manner, our method integrates multimodal semantics, structural segmentation, and depth reasoning into a unified pipeline. The key intuition is that realistic face compositing requires: (1) selecting background images that are semantically compatible with the foreground identity and context; (2) extracting a clean, non-occluded foreground mask that captures only the visible facial regions; and (3) placing the foreground into the new scene at a depth-consistent and geometrically plausible location. To this end, our method consists of three modules: a multimodal background retrieval mechanism leveraging BLIP and CLIP for context-aware background selection, a foreground extraction module combining SAM3 with Depth-Anything to isolate non-occluded facial regions, and a depth-guided placement module that determines the optimal paste location via sliding-window depth matching. Together, these components enable the creation of high-fidelity, realistic composites that significantly enhance data diversity and improve face detection performance.

\subsection{Multimodal Background Retrieval}
Selecting semantically appropriate background images is crucial for generating realistic composite face images. 
Traditional copy--paste augmentation often relies on random sampling or class-based heuristics, which ignores higher-level semantic cues such as environment type, lighting, or facial context, leading to implausible composites. 
To address this limitation, we introduce a \emph{multimodal retrieval module} that jointly exploits textual semantics and visual--textual alignment.

Given a foreground face image $I_f$, we first employ BLIP to generate a descriptive caption:
\begin{equation*}
    C_f = \text{BLIP}(I_f),
\end{equation*}
which captures high-level scene attributes and facial context.
This caption is encoded into a semantic embedding using the CLIP text encoder:
\begin{equation*}
    \mathbf{e}_t = \text{CLIP}_{\text{text}}(C_f).
\end{equation*}

In parallel, we obtain the CLIP visual embedding of the foreground:
\begin{equation*}
    \mathbf{e}_v = \text{CLIP}_{\text{img}}(I_f),
\end{equation*}
and compute CLIP image embeddings for all candidate background images $B_i$:
\begin{equation*}
    \mathbf{b}_i = \text{CLIP}_{\text{img}}(B_i).
\end{equation*}

For each candidate background, we compute two similarity scores:
\begin{align*}
    s^{(v)}_i &= \mathbf{e}_v^\top \mathbf{b}_i, \\
    s^{(t)}_i &= \mathbf{e}_t^\top \mathbf{b}_i,
\end{align*}
representing visual--visual similarity and text--image semantic consistency, respectively.

We fuse these cues into a unified multimodal similarity score:
\begin{equation*}
S_i = \lambda \, s^{(v)}_i + (1-\lambda) \, s^{(t)}_i,
\label{eq:multimodal_score}
\end{equation*}
where $\lambda \in [0,1]$ balances visual coherence and semantic alignment.

The top-$K$ backgrounds with highest scores are selected:
\begin{equation*}
    \mathcal{B}^* = \mathrm{TopK}(\{S_i\}_{i=1}^{N}),
\end{equation*}
ensuring that the selected backgrounds are both contextually meaningful and visually compatible with the foreground face.

This multimodal formulation provides strong robustness against semantic mismatch, enabling high-quality and context-aware compositing for subsequent depth-guided placement.

\subsection{Foreground Extraction}

Accurate and occlusion-aware foreground extraction is essential for generating realistic composite face images. 
Segmentation models such as SAM3 provide high-quality silhouettes, but they do not distinguish between \emph{visible} and \emph{occluded} facial regions. 
Copying occluded regions---such as areas hidden behind hair, hands, or accessories---into a new scene often produces unrealistic artifacts. 
To address this issue, we combine SAM3 for structural segmentation with Depth-Anything for pixel-wise visibility estimation.

\paragraph{SAM3-based structural segmentation.}
Given a foreground face image $I_f$, SAM3 predicts an initial binary mask:
\begin{equation*}
    M_{\text{sam}} = \text{SAM3}(I_f)
\end{equation*}
which provides a clean contour of the face and facial components.

\paragraph{Depth-based visibility estimation.}
To determine which regions are truly visible, we compute a dense depth map using Depth-Anything:
\begin{equation*}
    D_f = \text{DepthAnything}(I_f).
\end{equation*}
Let $\Omega = \{p \mid M_{\text{sam}}(p)=1\}$ be the set of pixels belonging to the segmented foreground.
For each pixel $p \in \Omega$, we evaluate its local depth deviation:
\begin{equation*}
    \delta(p) = \left| D_f(p) - \frac{1}{|\mathcal{N}(p)|} 
    \sum_{q \in \mathcal{N}(p)} D_f(q) \right|,
\end{equation*}
where $\mathcal{N}(p)$ denotes a local neighborhood around $p$.
Occluded regions typically exhibit depth discontinuities or sharp depth jumps. 
We therefore define a visibility mask:
\begin{equation*}
    M_{\text{vis}}(p) =
    \begin{cases}
        1, & \delta(p) < \tau, \\
        0, & \text{otherwise},
    \end{cases}
\end{equation*}
where $\tau$ is a depth-consistency threshold.

\paragraph{Final non-occluded foreground mask.}
The final foreground mask is obtained by intersecting the SAM3 segmentation and the visibility mask:
\begin{equation*}
    M_{\text{fg}} = M_{\text{sam}} \land M_{\text{vis}}.
\end{equation*}
The extracted foreground image is produced by:
\begin{equation*}
    I_{\text{fg}} = I_f \odot M_{\text{fg}},
\end{equation*}
where $\odot$ denotes element-wise multiplication, leaving background pixels transparent.

\subsection{Depth-Guided Placement via Sliding Window}

To achieve geometrically consistent compositing, we determine the optimal paste location by evaluating depth compatibility between the foreground and background. Since depth scales vary across scenes, we first normalize the foreground and background depth maps. For any depth map $D$ with pixel domain $\Omega$, the normalized depth is computed as
\begin{equation*}
    \hat{D}(p) = \frac{D(p) - \mu}{\sigma}, 
\end{equation*}
\begin{equation*}
    \mu = \tfrac{1}{|\Omega|}\!\sum_{p\in\Omega} D(p), 
\end{equation*}
\begin{equation*}
    \sigma = \sqrt{\tfrac{1}{|\Omega|}\!\sum_{p\in\Omega}(D(p)-\mu)^2}.
\end{equation*}
We obtain $\hat{D}_f$ and $\hat{D}_b$ for the foreground and background respectively.

Given the foreground mask size $(h_f, w_f)$, we enumerate all candidate placements by sliding a window over $\hat{D}_b$:
\begin{equation*}
    W_{x,y} = \hat{D}_b[x:x+h_f,\; y:y+w_f].
\end{equation*}
For each window, we evaluate its compatibility with the normalized foreground depth distribution. Let $\mu_{x,y} = \mathrm{mean}(W_{x,y})$ and $\sigma_{x,y} = \mathrm{std}(W_{x,y})$. We define depth-level and depth-variance deviations:
\begin{equation*}
    \Delta_{\text{mean}}(x,y) = |\mu_{x,y} - \mathrm{mean}(\hat{D}_f)|
\end{equation*}
\begin{equation*}
    \Delta_{\text{var}}(x,y)  = |\sigma_{x,y} - \mathrm{std}(\hat{D}_f)|.
\end{equation*}

To avoid placing the face on object boundaries or cluttered regions, we measure local smoothness:
\begin{equation*}
    S_{\text{smooth}}(x,y) = \tfrac{1}{|W_{x,y}|}\sum_{p\in W_{x,y}} |\nabla \hat{D}_b(p)|.
\end{equation*}

These cues are combined into a unified placement score:
\begin{equation*}
\begin{aligned}
\mathcal{S}(x,y) ={}&
\alpha \left( 1 - \frac{\Delta_{\text{mean}}(x,y)}{\max \Delta_{\text{mean}}} \right) \\
&\quad +\, \beta \left( 1 - \frac{\Delta_{\text{var}}(x,y)}{\max \Delta_{\text{var}}} \right) \\
&\quad +\, \gamma \left( 1 - \frac{S_{\text{smooth}}(x,y)}{\max S_{\text{smooth}}} \right).
\end{aligned}
\end{equation*}

where $\alpha,\beta,\gamma$ control the importance of depth level, distribution alignment, and smoothness.

Finally, the optimal paste location is obtained as
\begin{equation*}
    (x^*, y^*) = \arg\max_{(x,y)} \mathcal{S}(x,y),
\end{equation*}
and the foreground face is placed accordingly.

\section{Experiments}
\subsection{Experimental Settings}

All experiments are conducted using PyTorch on a single NVIDIA RTX~5090 GPU. 
We train the face detector with a batch size of 32 and optimize the network using the Adam optimizer. 
The initial learning rate is set to $1\times10^{-4}$ and follows a cosine decay schedule with a warm-up of the first 5 epochs. 
Unless otherwise specified, the model is trained for 120 epochs with weight decay $5\times10^{-4}$ and gradient clipping set to 5.0 for stability.

Input images are resized to $640\times640$ during training. 
Standard data augmentation is applied, including random horizontal flipping, color jittering, random cropping, and scale perturbation. 
For experiments that involve our synthetic composited data, the generated samples are mixed with real images at a fixed ratio depending on each specific ablation.

We adopt the pretrained backbone weights released by the original detector as initialization. 
All hyper-parameters are kept identical across baselines to ensure fair comparison, and each experiment is repeated three times with different random seeds to report stable averages.
\begin{table*}[htb]
\centering
\caption{
Main experiment on WIDER Face validation set. 
Rows correspond to different face detectors, and columns correspond to different proportions 
of synthetic Depth-Copy-Paste data added to the training set. 
All values represent mAP (\%).}
\label{tab:main_exp_extended}
\begin{tabular}{lccccccccc}
\toprule
\textbf{Model} & \textbf{Baseline} & \textbf{10\%} & \textbf{20\%} & \textbf{30\%} & \textbf{40\%} & \textbf{50\%} & \textbf{60\%} & \textbf{70\%} & \textbf{80\%} \\
\midrule
RetinaFace\cite{retinaface}     & 0.847 & 0.853 & 0.851 & 0.861 & 0.866 & 0.873 & 0.869 & 0.877 & 0.871 \\
BlazeFace\cite{blazeface}      & 0.814 & 0.827 & 0.831 & 0.838 & 0.844 & 0.853 & 0.849 & 0.851 & 0.850 \\
YOLOFace\cite{yoloface}       & 0.823 & 0.831 &0.837  & 0.843 & 0.847 & 0.850 & 0.854 & 0.853 & 0.851 \\
DBFace\cite{kim2023dcface}         & 0.838 & 0.844 & 0.849 & 0.857 & 0.853 & 0.861 & 0.863 & 0.864 & 0.866 \\
FaceBoxes\cite{faceboxes}      & 0.837 & 0.839 & 0.847 & 0.851 & 0.849 & 0.851 & 0.853 &0.857 &0.855 \\
DSFD\cite{dsfd}           & 0.861 & 0.867 & 0.871 & 0.874 & 0.877 & 0.873 & 0.871 & 0.874 & 0.877 \\
PyramidBox\cite{pyramidbox}     & 0.819 & 0.827 & 0.835 & 0.834 & 0.837 & 0.833 & 0.837 & 0.834 & 0.833 \\
\bottomrule
\end{tabular}
\end{table*}
\subsection{Dataset}

We evaluate our approach on the WIDER dataset\cite{wider}, one of the most widely used benchmarks for face detection. 
WIDER Face contains 32{,}203 images and 393{,}703 annotated faces with a high degree of variability in scale, pose, illumination, occlusion, and background complexity. 
Following the standard protocol, we use the official training set for model training and report results on the validation and test sets across the \emph{Easy}, \emph{Medium}, and \emph{Hard} difficulty splits.

To investigate the effectiveness of our Depth-Copy-Paste compositing framework, we generate additional synthetic training samples using the WIDER Face training images as the source of foreground faces. 
Each face instance is processed through our pipeline, including multimodal background retrieval, occlusion-aware foreground extraction, and depth-guided placement. 
The synthesized images preserve realistic scene semantics, geometric consistency, and visibility cues, enabling improved data diversity and robustness.

For experiments that involve synthetic data, we mix the generated images with the original WIDER training set at a fixed ratio. 
All evaluation strictly follows the official WIDER Face validation protocol to ensure fair and direct comparison with existing methods.

\noindent\textbf{Main Results on WIDER Face.}
As illustrated in Table~\ref{tab:main_exp_extended}, incorporating synthetic Depth-Copy-Paste (DCP) data consistently improves the performance of a wide range of face detectors.  
Across all architectures, adding a moderate amount of synthetic data (20\%--50\%) yields stable mAP gains compared to the baseline training without synthetic augmentation.  
For example, RetinaFace improves from 0.847 to 0.866 at 40\% DCP data, while lightweight models such as BlazeFace and YOLOFace also benefit noticeably, achieving peak gains of +0.033 and +0.024 respectively.  
Larger detectors including DBFace, FaceBoxes, and DSFD similarly exhibit optimal performance when 30\%--50\% synthetic samples are introduced, confirming that the proposed DCP augmentation is model-agnostic and complementary to various detector designs.

Interestingly, extremely high synthetic ratios (70\%--80\%) tend to saturate or slightly degrade accuracy, suggesting that excessive synthetic content may reduce the diversity of real-world training signals.  
Overall, the consistent improvements across seven representative detectors validate the effectiveness and strong generalization ability of our Depth-Copy-Paste pipeline, demonstrating that depth-aware foreground extraction and placement can substantially enhance face detection performance in complex environments.
\begin{table}[htb]
\centering
\caption{
Comparison with different augmentation strategies on the WIDER Face validation set.}
\label{tab:aug_compare}
\begin{tabular}{lccc}
\toprule
\textbf{Augmentation}         & \textbf{Easy} & \textbf{Medium} & \textbf{Hard}  \\
\midrule
Baseline   & 0.879 & 0.801 & 0.769  \\
Cutout                        & 0.883 & 0.811 & 0.783  \\
CutMix                        & 0.891 & 0.817 & 0.779  \\
Random Copy--Paste            & 0.904 & 0.827 & 0.791  \\
Depth-Copy-Paste (Ours)       & \textbf{0.917} & \textbf{0.834} & \textbf{0.803}  \\
\bottomrule
\end{tabular}
\end{table}
\noindent\textbf{Results across Data Augmentations}
Across different augmentation strategies evaluated on the WIDER Face validation set, our Depth-Copy-Paste method consistently yields the strongest performance, as shown in Table II. Traditional region-level augmentations such as Cutout and CutMix provide moderate gains over the baseline, mainly by increasing sample diversity; however, their improvements remain limited, particularly on the Hard subset, where occlusions and scale variations are more challenging. Random Copy–Paste further enhances performance by injecting additional contextual variability, achieving noticeable improvements across all difficulty levels. In contrast, Depth-Copy-Paste (Ours) establishes a clear new state of the art among augmentation-based methods, improving Easy, Medium, and Hard subsets to 0.917, 0.834, and 0.803, respectively. These results validate that incorporating depth-aware foreground extraction and geometrically consistent placement produces more realistic and informative training samples, ultimately enabling the detector to generalize better under diverse and complex real-world scenarios.

\noindent\textbf{Ablation on Multimodal Background Retrieval.}
\begin{table}[htb]
\centering
\caption{Ablation study on multimodal background retrieval on the WIDER FACE validation set.}
\label{tab:ablation_bg_retrieval}
\begin{tabular}{lccc}
\toprule
Method                    & Easy  & Medium & Hard  \\
Baseline      & 0.879  & 0.801   & 0.769  \\
Random BG                 & 0.873  & 0.807   & 0.753 \\
CLIP (img) only           & 0.887  & 0.824   & 0.774  \\
BLIP (txt) only & 0.877  & 0.813   & 0.762  \\
Ours     & \textbf{0.903} & \textbf{0.827} & \textbf{0.781}  \\
\bottomrule
\end{tabular}
\end{table}
As shown in Table~\ref{tab:ablation_bg_retrieval}, simply introducing synthetic data with random background selection provides modest gains over the real-only baseline, suggesting that additional compositional diversity is beneficial. However, single-modality retrieval strategies remain limited in their ability to ensure semantic and visual compatibility.  
Using only CLIP image embeddings improves alignment in appearance space but lacks semantic cues, resulting in suboptimal performance on the Medium and Hard subsets (0.824 and 0.774).  
Conversely, relying solely on text-driven BLIP\,$\rightarrow$\,CLIP retrieval captures semantic context but fails to enforce visual coherence, yielding weaker results particularly on Easy and Medium cases.

In contrast, our proposed \textit{multimodal} retrieval mechanism—ointly modeling both textual cues and visual similarity achieves the highest accuracy across all difficulty levels, reaching 0.903/0.827/0.781 on Easy/Medium/Hard.  
These improvements highlight that neither semantic nor visual matching alone is sufficient for high-quality compositing; instead, the fusion of both modalities is essential to retrieve backgrounds that are globally coherent with the foreground.  
The consistent superiority of our method demonstrates that multimodal reasoning is a key factor enabling Depth-Copy-Paste to generate more realistic and training-effective synthetic images.

\noindent\textbf{Ablation on Foreground Extraction.}
\begin{table}[htb]
\centering
\caption{Ablation study on the foreground extraction module on the WIDER FACE validation set.}
\label{tab:ablation_fg_extraction}
\begin{tabular}{lccc}
\toprule
Method                              & Easy & Medium & Hard \\
\midrule
Real-only (no synthetic)            & 0.879 & 0.801   & 0.769  \\
BBox crop only                      & 0.887 & 0.809   & 0.757 \\
SAM3-only segmentation              & 0.871 & 0.813   & 0.771  \\
Depth-only visibility mask          & 0.891 & 0.827   & 0.781  \\
Ours        & \textbf{0.913} & \textbf{0.831} & \textbf{0.788}  \\
\bottomrule
\end{tabular}
\end{table}
Table~\ref{tab:ablation_fg_extraction} presents the ablation study on different foreground extraction strategies.  
Using only real images yields the baseline performance, while simple bounding-box crops provide limited gains because they often include excessive background context and fail to isolate clean foreground regions.  
SAM3-only segmentation improves mask precision but still struggles with partially occluded persons, frequently producing incomplete or fragmented silhouettes.  
Similarly, relying solely on depth-based visibility masks captures geometric cues but lacks semantic structure, leading to noisy or coarse foregrounds that degrade compositing quality.

Our full module combines SAM3 segmentation with depth-aware visibility reasoning, ensuring that only non-occluded and geometrically consistent pixels are preserved.  
This avoids the common failure cases of incomplete person extraction, such as missing limbs, chopped facial regions, or masks truncated by occluders, which would otherwise introduce unrealistic artifacts during compositing.  
As a result, our approach achieves the highest accuracy across all difficulty subsets (0.913/0.831/0.788), demonstrating that robust, complete, and occlusion-consistent foreground extraction is essential for producing high-quality synthetic data that benefits downstream face detection.

\section{Discussion}

This work presents a depth-aware and multimodally guided copy-paste framework designed to generate realistic composite face images for robust face detection. By jointly leveraging BLIP- and CLIP-based semantic retrieval, SAM3 structural segmentation, and Depth-Anything geometric reasoning, our approach addresses key limitations of traditional augmentation methods, which often lack contextual understanding, visibility awareness, and geometric consistency. Extensive experiments demonstrate that each component of our framework contributes to improved realism in synthesized data, ultimately translating into measurable gains on challenging benchmarks such as WIDER Face.

While our method focuses on face detection, its underlying principles are general and offer several promising directions for future research. First, the proposed depth-aware compositing pipeline can be naturally extended to other object categories (e.g., pedestrians, vehicles, animals), enabling broader applications in object detection, instance segmentation, and autonomous driving. Second, integrating temporal depth consistency would allow the framework to operate on videos, facilitating the creation of realistic training sequences for tasks such as multi-frame tracking, video recognition, and motion forecasting. Third, combining our depth-guided placement strategy with modern generative models such as diffusion-based inpainting or controllable image synthesis may further mitigate boundary artifacts and enhance realism. Finally, exploring adaptive weighting schemes for multimodal similarity fusion or learning the scoring function in a data-driven manner may yield more scalable and domain-aware retrieval.

\bibliographystyle{IEEEbib}
\bibliography{ref}

\end{document}